\documentclass[accepted]{uai2025} 
                        
\usepackage[british]{babel}

\usepackage{natbib} 
\usepackage{mathtools} 
\usepackage{microtype}
\usepackage{graphicx}
\usepackage{subfigure}
\usepackage{booktabs} 
\usepackage{multirow}
\usepackage{enumitem}
\usepackage{tikz} 
\usepackage{amsmath}
\usepackage{amssymb}
\usepackage{mathtools}
\usepackage{amsthm}
\usepackage{bm}

\usepackage{tabularx}
\usepackage{algorithm}
\usepackage{algorithmic}
\newcommand{\leftcomment}[1]{\textit{\# #1}}
\usepackage{xcolor}
\usepackage{mdframed}
\usepackage{tcolorbox}

\theoremstyle{plain}
\newtheorem{theorem}{Theorem}[section]

\theoremstyle{definition}
\newtheorem{definition}[theorem]{Definition}

\theoremstyle{remark}

\title{A Unified Data Representation Learning for \\ Non-parametric Two-sample Testing}

\author[1]{Xunye Tian}
\author[2]{Liuhua Peng}
\author[1]{Zhijian Zhou}
\author[2]{Mingming Gong}
\author[3]{Arthur Gretton}
\author[1]{Feng Liu}
\affil[1]{%
    Faculty of Engineering and Information Technology, University of Melbourne, AU
}
\affil[2]{%
    School of Mathematics and Statistics, University of Melbourne, AU
}
\affil[3]{%
    Gatsby Computational Neuroscience Unit, University College London, UK
}
  
\begin{document}
\maketitle

\begin{abstract}
    Learning effective data representations has been crucial in non-parametric two-sample testing. Common approaches will first split data into training and test sets and then learn data representations \emph{purely on the training set}. 
    However, recent theoretical studies have shown that, as long as the sample indexes are not used during the learning process, the \emph{whole data} can be used to learn data representations, meanwhile ensuring control of Type-I errors.
    The above fact motivates us to use the test set (but \emph{without} sample indexes) to facilitate the data representation learning in the testing.
    To this end, we propose a \emph{representation-learning two-sample testing} (RL-TST) framework. RL-TST first performs purely self-supervised representation learning on the entire dataset to capture \emph{inherent representations} (IRs) that reflect the underlying data manifold. A discriminative model is then trained on these IRs to learn \emph{discriminative representations} (DRs), enabling the framework to leverage both the rich structural information from IRs and the discriminative power of DRs. Extensive experiments demonstrate that RL-TST outperforms representative approaches by \emph{simultaneously} using data manifold information in the test set and enhancing test power via finding the DRs with the training set.
\end{abstract}

\section{Introduction}\label{sec:intro}
Two-sample tests aim to answer a question: ``Are two samples drawn from the same distribution?''. Classical two-sample tests, including \textit{t}-tests which test the empirical mean differences between two samples, often need to assume that samples are drawn from specific distributions (e.g., Gaussian distributions with the same variance). To alleviate the assumptions, non-parametric two-sample tests are proposed to solve the problem only based on observed data 
\citep{GSSSP12:kernel-choice,Heller2016,Szekely2013,Jitkrittum2016,Chen2017,Ghoshdastidar2017,Lopez:C2ST,RamdasGarciaCuturi,sutherland:mmd-opt,Gao18neurips,Graph_two_sample,Lerasle2019monk,liu2020learning,Kirchler2020two,Kubler2020learning,cheng2021neural,jonas2022awit,jonas2022automl,liu_meta_2021,Deka2023mmd,Bonnier2023kernel,schrab2023mmd, biggs_mmd-fuse_2023}.

For example, the \emph{Kolmogorov-Smirnov} (K-S) test is designed to compare the {cumulative distribution functions} derived from two samples, but generalisation to higher dimension is challenging \citep{bickel1969distribution}. The \emph{maximum mean discrepancy} (MMD) test adopts the kernel mean embedding of distribution and uses it to measure the discrepancy between two distributions \cite{GBRSS12:mmd} whose dimensions can be relatively higher than classical methods \citep{liu2020learning}. The statistics used in these non-parametric two-sample tests are also widely adopted in many other fields, such as domain adaptation, causal discovery, generative modeling, adversarial learning, and more \citep{Gong2016,MMD_GAN,DA_app_Stojanov,cano2020kappa,ODLCMP20:fair-reps,gao2021max,fang2021open,zhong2021how,fang2021learning,song2021segment,tahmasbi2021driftsurf,tasksen2021sequential,Bergamin2022model}.

To improve the test power of non-parametric two-sample tests in practical applications, recent studies have shown that learning good data representations is crucial before performing two-sample testing \citep{Kirchler2020two,liu2020learning,liu_meta_2021,gao2021max,Bergamin2022model}. For example, \citet{Kirchler2020two} directly use a pre-trained feature extractor to extract features of two samples and find it is useful to increase the test power during the testing. Meanwhile, \citet{liu2020learning} propose a learning paradigm to learn deep-net representations of data via maximizing the test power of MMD and show that the learned representations can help capture the difference between two complex-structured samples. Even though utilizing a fraction of the samples to train a classifier \citep{Lopez:C2ST} or a kernel function \citep{liu2020learning} enables deriving \emph{discriminative representations} (DRs) of the remaining samples, the data splitting process results in a \emph{trade-off} between the extra power provided by the learned functions/kernels and the sacrificed power due to the decreasing testing sample size.

However, \citet{biggs_mmd-fuse_2023} have pointed out that, after discarding the sample information (namely, we do not know which sample the data belongs to), learning purely \emph{inherent representations} (IRs) from aggregated samples will not influence the type I error of permutation-based testing methods, but the missing discriminative power makes it underperform on the complex-structured data, which can further justify the correctness of learning good representations for testing. 


Motivated by the above theoretical studies and existing challenges, we propose a \emph{representation-learning two-sample testing} (RL-TST) framework that focuses on learning good representations on the samples, from both IRs and DRs. Since two-sample testing data mainly follows a manifold assumption where the (high-dimensional) data lie (roughly) on a low-dimensional manifold, we could firstly learn an encoder from the representation learning that is responsible for extracting the IRs of \emph{entire samples}. Then, train a discriminative model on the learned IRs will enable the model with discriminative ability directly on the inherent manifolds of samples rather than on the complex embedded space of samples. 
This framework captures the sample structure information discarded in the data splitting process and exhibits a higher discriminative power than purely unsupervised representation learning on the entire dataset.

We conduct extensive experiments to implement RL-TST on different kinds of MMD-based two-sample testing methods, we verify the empirical effectiveness of RL-TST over the \emph{state-of-the-art} (SOTA) two-sample testing methods on synthetic \emph{high-dimensional Gaussian mixture} (HDGM) dataset, MNIST dataset and ImageNet dataset. These are the commonly used benchmarks to detect the performance of two-sample testing methods. 

Our main contributions are:
\begin{itemize}
    \item We propose a novel RL-TST, which can address the challenges of two existing frameworks and provide a new research direction of two-sample testing. 
    \item Empirically, the various implementations of RL-TST show their outperformances over different SOTA baselines across different benchmarks.
    \item Comparatively, we provide the discussion and empirical evidence of why alternative frameworks, such as semi-supervised learning or purely self-supervised learning are facing challenges in two-sample testing scenarios. 
\end{itemize}

\section{Preliminaries} \label{preliminary}

\textbf{Two-sample Testing.}
Two-sample testing is one of the statistical hypothesis tests that aims to assess whether two \emph{independent and identically distributed} samples, denoted by \(S_\mathbb{P} = \{x_i\}_{i=1}^n \sim \mathbb{P}^n\) and \(S_\mathbb{Q} = \{y_j\}_{j=1}^m \sim \mathbb{Q}^m\), where $x_i, y_j \in \mathcal{X}$, are drawn from the same distribution \citep{lehmann2005testing}. In two-sample testing, the \textit{null hypothesis} $H_0$ refers to two samples being drawn from the same distribution, which corresponds to $\mathbb{P} = \mathbb{Q}$. The \textit{alternative hypothesis} $H_1$ indicates that two samples are drawn from different distributions, meaning $\mathbb{P} \neq \mathbb{Q}$. Whether we should accept or reject $H_0$ depends on the test statistic $\hat{t}$, which represents the differences between two samples. 


\textbf{Classifier Two-sample Testing (C2ST).}
C2ST aims to train a binary classifier: if the classifier obtains a testing accuracy significantly better than random guessing, it suggests that the two samples come from different distributions \citep{Lopez:C2ST}. Specifically, given dataset $\mathcal{S} = \{(x_i, 0) | x_i \in S_\mathbb{P}\}_{i=1}^{n} \cup \{(y_j, 1) | y_j \in S_\mathbb{Q}\}_{j=1}^{m} := \{(z_k, l_k)\}_{k=1}^{m+n}$, where $m = n$, and we can shuffle and split $\mathcal{S}$ into training set $\mathcal{S}_{\rm tr}$ and testing set $\mathcal{S}_{\rm te}$, let $f^* : \mathcal{X} \rightarrow \{0, 1\}$ be a binary classifier that is well-trained on $\mathcal{S}_{\rm tr}$, then the test statistic or the accuracy of the classifier $f^*$ on $\mathcal{S}_{\rm te}$ can be written as:
\begin{equation}
    \label{c2st_stats}
        \hat{t} = \frac{1}{n_{\rm te}}\sum_{(z_k, l_k) \in \mathcal{S}_{\rm te}} \mathbb{I} \left[ f^*(z_k) = l_k \right],
\end{equation}
where $n_{\rm te} = |\mathcal{S}_{\rm te}|$ and $\mathbb{I}$ is the indicator function. Finally, we compute the $p$-value to determine if the test statistic is significantly greater than the random guessing accuracy, utilizing the approximate null distribution of C2ST \citep{Lopez:C2ST, kim2020classificationaccuracyproxysample}
and the permutation test \citep{good_permutation}.

\textbf{C2ST with logits (C2ST-L).}
Moreover, we can also consider using the trained classifier $f^*$ in C2ST not directly to compute the accuracy but to extract representations of two samples \citep{cheng2020classificationlogittwosampletesting}. Let $h$ be the feature extractor of $f^*$, then $h(z)$ (could be the model's output, i.e., logit, or the model's hidden-layer output) can be regarded as representations of two samples as the new two samples. For these new two samples, we can use MMD (with a linear kernel) to compute the difference between two samples. Let $S_\mathbb{P}^{\rm te}$ and $S_\mathbb{Q}^{\rm te}$ be the splitting samples of $S_\mathbb{P}$ and $S_\mathbb{Q}$ in the testing set $\mathcal{S}_{\rm te}$ and $n_{x}^{\rm te}$ and $n_{y}^{\rm te}$ be the sample size of $S_\mathbb{P}^{\rm te}$ and $S_\mathbb{Q}^{\rm te}$. In general, the statistic used in C2ST-L is
\begin{equation}
    \label{c2st_l_stats}
    \hat{t}_{\mathit{L}} = \left\|\frac{1}{n_{x}^{\rm te}}\sum_{x_i\in S_\mathbb{P}^{\rm te}}h(x_i) - \frac{1}{n_{y}^{\rm te}}\sum_{y_i\in S_\mathbb{Q}^{\rm te}}h(y_i)\right\|_{2}^2,
\end{equation}
where $\|\cdot\|_2$ is the L2 norm. When $h(z)$ is logits, C2ST-L is the same as C2ST-L used by \citet{liu2020learning}.

\textbf{Maximum Mean Discrepancy (MMD) Test with Deep Kernel (MMD-D).} A quick recap on unbiased $U$-statisticd estimator for $\rm MMD^2$ when $m = n$:
\begin{equation}
    \label{mmd_stats}
    \widehat{\rm MMD}^2_u(S_\mathbb{P}, S_\mathbb{Q}; k) := \frac{1}{n(n-1)}\sum_{i\neq j}H_{ij}
\end{equation}
\[
    H_{ij} := k(x_i, x_j) + k(y_i, y_j) - k(x_i, y_j) - k(x_j, y_i).
\]
Compared to training a classifier, MMD-D focuses on learning a powerful deep kernel function $k_\theta$, 
\begin{equation}\label{eq:deepkernel}
    k_\phi(x, y) = [(1-\epsilon)\kappa(\phi(x), \phi(y)) + \epsilon]q(x, y), 
\end{equation}
where $\phi : \mathcal{X} \rightarrow \mathbb{R}^k$ is the deep neural network (with parameters $\theta_\phi$) which outputs the DRs of samples, $\epsilon$ is the interpolation weigth that $0 < \epsilon < 1$, and $\kappa$ and $q$ are characteristic kernels with hyperparameters $\theta_\kappa$ and $\theta_q$ respectively.
To ensure the deep kernel can directly measure the distance of representations of complex-structured samples, optimizing the kernel with the highest test power will approximately maximize  \citep{sutherland:mmd-opt, liu2020learning}
\begin{equation}
    \label{mmd_objective}
    \mathcal{J} := {\rm MMD}^2(\mathbb{P}, \mathbb{Q}; k_\phi) / \sigma_{\mathcal{H}_1}(\mathbb{P}, \mathbb{Q}; k_\phi),
\end{equation}
where $\sigma^2_{\mathcal{H}_1}(\mathbb{P}, \mathbb{Q}; k_\phi) := 4(\mathbb{E}[H_{12}H_{13}] - \mathbb{E}[H_{12}]^2])$ is the variance of $\sqrt{n}{\rm MMD}^2_u - {\rm MMD}^2$ under the alternative hypothesis $H_1: \mathbb{P} \neq \mathbb{Q}$ by a standard central limit theorem \citep{liu2020learning}, and the $H_{ij}$ follows the definition above.  

\textbf{Permutation Testing.} According to the standard central limit theorem \citep{serfling2009approximation}, the test statistic $\hat{t}$ in~\eqref{c2st_stats} converges to normal distributions under both the null or alternative hypothesis \citep{Lopez:C2ST}. Although it is feasible for us to derive the threshold $t_{\alpha}$ of the null hypothesis distribution and perform a traditional Z-Test, it is simpler and faster to instead implement a permutation test for all test statistics~\eqref{c2st_stats}, ~\eqref{c2st_l_stats} and ~\eqref{mmd_stats} \citep{sutherland:mmd-opt}. We will permute and randomly assign samples to new $S_\mathbb{P}^{\rm te'}$ and $S_\mathbb{Q}^{\rm te'}$ for $n$ times. Under $H_0$, the samples from $\mathbb{P}$ and $\mathbb{Q}$ should be interchangeable, implying that the test statistic should exhibit minimal variation between its value based on the original sequence of samples and its computation from several randomly permuted sequences. Thus, if the original test statistic is large enough than most of the statistic derived from the randomly permuted sequences, we can conclude that we reject $H_0$ \citep{good_permutation}.

\begin{figure*}[ht]
\begin{center}
\centerline{\includegraphics[width=1.6\columnwidth]{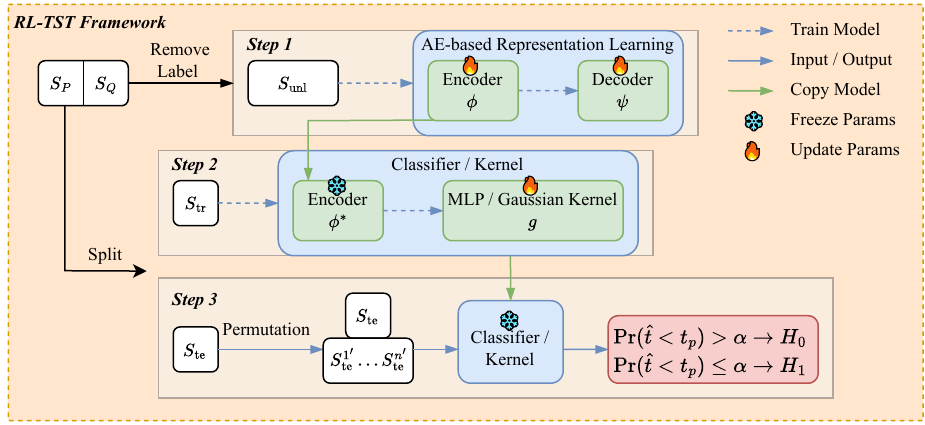}}
\caption{Overview of the RL-TST framework. Firstly, an encoder was learned from any AE-based representation learning algorithms on whole data, which can chosen from standard auto-encoder, wasserstein auto-encoder, etc. Secondly, fine-tune the learned encoder followed by a component that has the discriminative ability. At last, utilizing the final classifier or deep kernel to perform the permutation test based on statistic \eqref{c2st_stats}, \eqref{c2st_l_stats} or \eqref{mmd_stats} to derive the two-sample testing result.}
\label{model_arch}
\end{center}
\vspace{-2.2em}
\end{figure*}

\section{Representation-learning Two-sample Testing}
In this section, we introduce our proposed RL-TST framework and several implementations that could leverage the information from unlabelled data in two-sample testing. Next, we provide an understanding of why learning good representations could enhance the power of two-sample testing. At the end, we discuss the significant challenges if we want to use mainstream semi-supervised learning methods (e.g., methods based on label propagation \citep{lee2013pseudo}) to address two-sample testing problems, which is another framework to exploit information from unlabelled data. 

\subsection{Our Proposal: RL-TST}
\label{proposal}
Due to the unique properties of two-sample testing data, two samples often follow two very similar but in fact different distributions under the alternative hypothesis, which makes it impossible to obtain effective information from unlabeled samples through most self-supervised label propagation \citep{lee2013pseudo} or augmentation-based \citep{grill2020bootstraplatentnewapproach} techniques. From the recent studies, \citet{cheng2024kerneltwosampletestsmanifold} claims that most of the two-sample testing data follow the manifold assumption, where the data are low-dimensional intrinsic manifolds embedded in high-dimensional space. Thus, we propose to use a two-phase pipeline in two-sample testing that leverages the labelled and unlabelled samples to learn IRs and DRs respectively \citep{dai2015semisupervisedsequencelearning}. 

Generally, since the effectiveness of \emph{auto-encoder-based} (AE-based) representation learning \emph{mainly relies on the manifold assumption} \citep{vincent2008extracting}, the first phase is an unsupervised AE-based representation learning, which learns a feature extractor that captures the inherent features for both samples. The next phase is to train a \emph{multilayer perceptron} (MLP, used to classify two samples) or a characteristic kernel (with optimized parameters) on those IRs of two-sample testing data, so the final model will exhibit the discriminative ability directly on the intrinsic manifolds of two samples \citep{belkin06a}. Finally, we apply the final model to the remaining samples (excluded in the second phase) to obtain their DRs and perform permutation testing on DRs to derive the final testing result. Overall, our framework can be generalised in three main steps: \emph{learn IRs, learn DRs, and then testing}. The general framework can be visualised in Figure \ref{model_arch}. In the following, we will introduce our framework in detail.


\textbf{Learning Details.} Since RL-TST has two phases, for C2ST, we need to decompose the classifier-based model $f$ into two parts: a feature extractor $\phi \in \mathcal{F}: \mathcal{X} \rightarrow \mathbb{R}^k$ that used to learn IRs and followed by a classifier $g \in \mathcal{G} : \mathbb{R}^k \rightarrow \{0, 1\}$ that used to learn DRs. We denote by $\phi_f$ and $g_f$ the feature extractor and the classifier of a specified model $f$. For the input samples, removing the label information will leave an unlabeled dataset $\mathcal{S}_{\rm unl} = \{z_k\}_{k=1}^{m+n}$ that is equal to $S_\mathbb{P} \cup S_\mathbb{Q}$.

\textbf{Learning IRs.} The first step is to train a representation learning encoder on the whole unlabelled dataset $\mathcal{S}_{\rm unl}$, with the training objective mainly to minimize the differences between input and reconstructed output. Generally, we aim to learn a featurizer $\phi^*$ such that
\begin{equation}
\label{IR}
    \phi^*, \psi^* = \arg\min_{\phi, \psi} \; \widehat{\mathcal{R}}_{\rm IR}(\phi, \psi),
\end{equation}
where $\psi : \mathbb{R}^k \rightarrow \mathcal{X}$ is the decoder. For a specific example (e.g., \emph{mean squared error} (MSE) in basic auto-encoder)
\begin{equation}
\label{IR_example}
    \widehat{\mathcal{R}}_{\rm IR}(\phi, \psi) = \frac{1}{|\mathcal{S}_{\rm unl}|}\sum_{z_i \sim \mathcal{S}_{\rm unl}} \left\| \psi(\phi(z_i)) - z_i \right\|_2^2,
\end{equation}
and the objective will be slightly varied with some penalization terms depending on different AE-based algorithms, such as variational auto-encoder \citep{Kingma_2019} or wasserstein auto-encoder \citep{tolstikhin2019wassersteinautoencoders}. After training, $\phi^*(z_i)$ is called the IR of $z_i$.

\textbf{Learning DRs.} Then, utilize the featurizer $\phi^*$ from the representation learning model and concatenate with either an MLP $g$ or a deep kernel $k$ to form a final model $\mathcal{M}$.
The combined model is fine-tuned on $\mathcal{S}_{\rm tr}$, focusing on maximizing the distance of MLP's output of two samples or the test power of MMD regarding the two samples. Formally, for MLP-based $\mathcal{M} := g \circ \phi^*$, we aim to learn a function $g^*$ on $\mathcal{S}_{\rm tr}$ by minimizing  
\begin{equation}
    \label{DR_clf}
    \mathcal{L}_{\rm DR}(g) = \frac{1}{|\mathcal{S}_{\rm tr}|} \sum_{(z_i, l_i) \sim \mathcal{S}_{\rm tr}} \ell_{\rm DR}(\phi^*(z_i),l_i, g),
\end{equation}
where $ \ell_{\rm DR}(\phi^*(z_i),l_i, g)$ can be empirically implemented using a loss function such as \emph{binary cross entropy} (BCE) loss, defined for binary classification as:
\begin{equation}
        \label{clf_objectives}
        \hat{\ell}_{\rm DR}(\phi^*(z_i),l_i, g) = 
        -[l_i\log \hat{p}_i + (1-l_i)(1-\log \hat{p}_i)],
\end{equation}
where $\hat{p}_i = \left(1 + g \circ \phi^*(z_i)\right)^{-1}$ is the estimate of $p$. As $g^*$ is an MLP, so $g^*$ can be expressed by $g^* = h^* \circ h^*_{\rm rep}$ where $h^*_{\rm rep}\in\{h_{\rm rep}: \mathbb{R}^k\rightarrow \mathbb{R}^{d_{\rm rep}}\}$ and $h^*\in\{h: \mathbb{R}^{d_{\rm rep}}\rightarrow \{0,1\}\}$. Normally, $h^*$ is called a classification head, and $h^*_{\rm rep}$ is called a representation function. Thus, a DR of $z_i$ is $h^*_{\rm rep} \circ \phi^*(z_i)$ if we use C2ST-based methods for testing.

For a MMD-based $\mathcal{M} := k_{\phi^*}$, we aim to empirically learn a deep kernel $k^*$ (shown in Eq.~\eqref{eq:deepkernel}) on the $\mathcal{S}_{\rm tr}$ by maximizing the empirical estimate of $\mathcal{J}$ in \eqref{mmd_objective}
\begin{equation}
    \label{DR_mmd}
    \widehat{\mathcal{J}}_{\rm DR}(S_\mathbb{P}^{\rm tr}, S_\mathbb{Q}^{\rm tr}; k_{\phi^*}) = \frac{\widehat{{\rm MMD}}_u^2(S_\mathbb{P}^{\rm tr}, S_\mathbb{Q}^{\rm tr}; k_{\phi^*})}{ \hat{\sigma}_{\mathcal{H}_1, \lambda} (S_\mathbb{P}^{\rm tr}, S_\mathbb{Q}^{\rm tr}; k_{\phi^*})}
\end{equation}
where $S_\mathbb{P}^{\rm tr}$ and $S_\mathbb{Q}^{\rm tr}$ be the splitting samples of $S_\mathbb{P}$ and $S_\mathbb{Q}$ in the training set $\mathcal{S}_{\rm tr}$ and $n_{x}^{\rm tr}$ and $n_{y}^{\rm tr}$ be the sample size of $S_\mathbb{P}^{\rm tr}$ and $S_\mathbb{Q}^{\rm tr}$. $\hat{\sigma}_{\mathcal{H}_1, \lambda}$ represents for the regularized estimator of $\sigma_{\mathcal{H}_1}$ defined in \citep{liu2020learning}.

\textbf{Testing.} 
In the end, compute any of the three test statistics in~\eqref{c2st_stats} (by setting $f^*$ as $g^* \circ \phi^*$), in \eqref{c2st_l_stats} (by setting $h^*$ as $h^*_{\rm rep} \circ \phi^*$, or in \eqref{mmd_stats} (by setting $k^*(\cdot, \cdot) = [(1-\epsilon)\kappa^*(\phi^*(\cdot), \phi^*(\cdot)) + \epsilon]q^*(\cdot, \cdot)$ based on the original sequence of samples and the $r$ times permuted samples, reject $H_0$ if original statistic is larger than the threshold derived from permuted statistics. The difference of these three statistics are pure accuracy in \eqref{c2st_stats}, linear kernel that contain confidence information from accuracy in \eqref{c2st_l_stats}, and higher-order deep kernel that explains complex structure information in \eqref{mmd_stats}.

\textbf{Discussion of Alternatives.} In the first phase, if the data sometimes follow a smoothness assumption where \emph{small perturbations will not influence the distribution of data} (i.e., the distance between two distributions will be significantly larger than the distance between the augmentations \citep{xie2020unsupervised}), an augmentation-based self-supervised representation learning can also be used \citep{grill2020bootstraplatentnewapproach, NEURIPS2021_83004190}. Instance-level representation learning through contrastive discrimination, like SimCLR \citep{chen2020simpleframeworkcontrastivelearning}, is not recommended in two-sample testing scenarios, even when both the smoothness and manifold assumptions are satisfied, since only holistic approaches, such as BYOL \citep{grill2020bootstraplatentnewapproach}, can effectively capture the IRs of whole data. 
However, we are proposing a general framework that can be applied to all the two-sample data, so we do not include such representation learning algorithm in the framework. 

\subsection{Understanding the Increased Test Power in General}\label{sec:understand}
In recent theoretical research of two-sample testing, as shown in Table \ref{theory_result_table}, researchers have found that the lower dimension is reduced relative to the sample size, the higher-order moment discrepancy the MMD test is capable to detect \citep{yan2023kernel}. Therefore, if we learn better representations of input samples under the manifold assumption, it will effectively extract the lower dimensional features without increasing sample size, making the MMD test more likely to capture the higher-order discrepancy. Since C2ST or C2ST-L are essentially MMD-based two-sample testing methods with sign kernel or linear kernel \citep{liu2020learning}, it is compelling that if we can learn better representations on the two-sample testing data, we will derive higher test power from any two-sample testing methods.

\begin{table}[!t]
\centering
\caption{Main theoretical results from \citep{yan2023kernel}, which displays that the relationship between the dimension of the data (denoted by $p$) and the sample size (denoted by N) will affect the $l$-order moment discrepancy the kernel two-sample testing being detected.}
\resizebox{0.5\textwidth}{!}{%
\begin{tabular}{@{}l|l@{}}
\toprule
\textbf{Dimension and sample size orders} & \textbf{Main features captured} \\ 
\midrule
$N = o(\sqrt{p})$                      & Mean and trace of covariance    \\ 
$N = o(p^{3/2})$                       & Mean and covariance             \\ 
$N = o(p^{l-1/2})$                     & The first $l$th moments         \\ 
Fixed $p$, growing $N$                 & Total homogeneity               \\ 
\bottomrule
\end{tabular}
}
\label{theory_result_table}
\end{table}


\subsection{Can We Use Semi-supervised Learning Methods for Testing?}
After introducing the RL-TST framework we proposed, which can make good use of the information from unlabelled data to improve test power. In machine learning, there is another modern technique called semi-supervised learning, which is also designed to utilize the information from unlabelled data to improve classification. In this section, we will discuss why mainstream \emph{semi-supervised learning} (SSL) methods (e.g., label propagation \citep{lee2013pseudo}) cannot be generally used to address two-sample testing problems where $S_{\rm tr}$ is regarded as the training set and $S_{\rm unl}$ is regarded as the unlabeled set.

We first recall the basic assumptions required by SSL methods \citep{mit06ssl}:


\begin{itemize}[leftmargin=2em, itemsep=0pt, topsep=0pt, parsep=0pt, partopsep=0pt]
    \item \textit{Smoothness assumption:} If points $x_1$ and $x_2$ are close, then so should be their labels $y_1$, $y_2$.  
    \item \textit{Cluster assumption:} If points are in the same cluster, they are likely to be of the same class.
    \item \textit{Manifold assumption:} The (high-dimensional) data lie (roughly) on a low-dimensional manifold.
\end{itemize}
Based on those assumptions, there are five representative semi-superivsed learning frameworks \citep{yang2023sslsurvey}: consistency-regularisation \citep{xie2020unsupervised}, pseudo-labelling \citep{lee2013pseudo}, graph-based \citep{song2021graphbased}, generative-models \citep{kingma2013auto} and hybrid \citep{sohn2020fixmatch} SSL methods. The consistency regularisation techniques assume that the model can predict the same label between the augmented or permuted samples and original samples; the pseudo-labelling techniques assume that if the samples form a cluster, then all of samples have same label in the same cluster; graph-based techniques assume that the input samples are graph-structured or the input can be represented as graph-structured data; the generative-model techniques assume that the generative samples have the same distribution as input samples. Hybrid techniques can embrace the advantages of the above techniques, however they also require all assumptions to hold. The details of above SSL methods are demonstrated in Appendix \ref{overview_sota_ssl}.

Even though those methods are comprehensive and advanced in the field of SSL to leverage the information from unlabelled data, they are inherently \emph{incompatible} to the two-sample testing scenarios. In general case, data in two-sample testing often form a high-degree of overlapping between two samples, which will decrease the useful information content of unlabelled data \citep{mit06ssl}, and the empirical verification on the challenges of applying semi-supervised techniques is presented below.

\begin{figure*}[]
    \centering
    \includegraphics[width=0.85\textwidth]{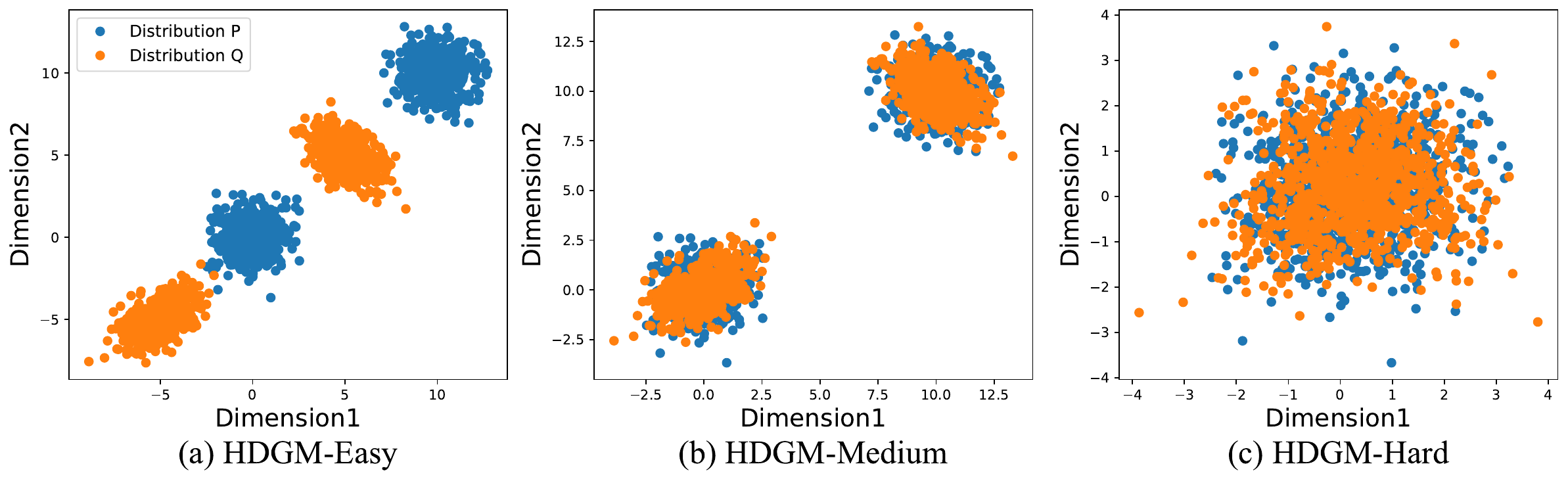}
    \caption{\small Visualisation of first two dimensions of samples for different levels of the high-dimensional Gaussian mixture (HDGM) dataset whose dimension is 10. For the HDGM-Easy and HDGM-Medium, the cluster mean difference $\Delta_{\mu}$ within the same distribution is 10, while for the HDGM-Hard, $\Delta_{\mu}$ is 0.5. For the HDGM-Easy, the distribution mean difference $\Delta_{q}$ between $\mathbb{P}$ and $\mathbb{Q}$ is 5, while for HDGM-Medium and HDGM-Hard, $\Delta_{q}$ is 0. Other setting of how to generate HDGM dataset is described in Appendix \ref{hdgm_dataset_detail}.}
    \label{HDGM_overview}
\end{figure*}

\textbf{Empirical Results for Validity of Mainstream Semi-supervised Learning Techniques.} In Figure \ref{HDGM_overview}, it shows different levels of overlap in the two-sample testing data, and we will conduct experiments on these datasets. Since SSL methods mainly applied on the classifier-based model, we explore the performance of C2ST-based methods. 

In Table \ref{SOTA-ssl-result}\footnote{The experimental details of this table can be found in Appendix \ref{overview_sota_ssl}, where all detailed description of semi-supervised methods and how to use these methods in testing are introduced.}, 
the empirical results show that even though the application of mainstream SSL methods on C2ST can have better performance on the HDGM-Easy dataset,
\begin{table*}[ht]
\centering
\small
\caption{\small Result of C2ST test power on HDGM-Easy, HDGM-Medium and HDGM-Hard (d=10), on different total size of two samples $N$ inputed in 100 trials. Compared to other application of mainstream SSL methods on C2ST, where C2ST-CR, C2ST-PL, C2ST-GM, and C2ST-HB represent that we learn the classifier of C2ST using four different mainstream SSL frameworks.
\protect\footnotemark}
\label{SOTA-ssl-result}
\begin{tabular}{@{}llllllllll@{}}
\toprule
    \multicolumn{1}{c}{\multirow{2}{*}{Method}}     & \multicolumn{3}{c}{HDGM-Easy} & \multicolumn{3}{c}{HDGM-Medium} & \multicolumn{3}{c}{HDGM-Hard} \\ 
          \cmidrule(l){2-4} \cmidrule(l){5-7} \cmidrule(l){8-10}
   & N=60   & N=80   & N=100   & N=2000    & N=3000   & N=4000   & N=4000   & N=6000   & N=8000  \\
         \cmidrule(r){1-1} \cmidrule(l){2-4} \cmidrule(l){5-7} \cmidrule(l){8-10}
C2ST     &    0.64     &    0.91      &    0.99      &    \textbf{0.44}       &     0.82     &   \textbf{0.97}       &    0.29      &   \textbf{0.49}      &    \textbf{0.78}      \\
\midrule
C2ST-CR  &    0.65     &    0.92      &    \textbf{1.00}      &    0.40       &     0.84     &   \textbf{0.97}       &    0.32      &   0.42      &   0.75       \\
C2ST-PL  &    0.72     &    0.96      &    0.99      &    0.40       &     0.76     &   0.93       &    \textbf{0.36}      &    0.45     &    0.77      \\
C2ST-GM  &    0.64     &    0.92      &    \textbf{1.00}      &    0.43       &     \textbf{0.85}     &   \textbf{0.97}       &    0.22      &    0.40     &    0.72      \\
C2ST-HB  &    \textbf{0.99}     &    \textbf{1.00}      &    \textbf{1.00}     &    0.25       &     0.43     &   0.58       &   0.28       &   0.43      &     0.65     \\ 
\bottomrule
\end{tabular}%
\vspace{-1em}
\end{table*}
but it often yields poorer results compared to the original C2ST on HDGM-Medium and HDGM-Hard datasets, which represents the common overlapping distribution data in the context of two-sample testing. This underperformance can be attributed to the fundamental nature of the testing procedure, which is distinct from accuracy evaluation in the classification tasks. In two-sample testing, our aim is to maximize the distance of two whole samples, rather than focusing on correctly classifying all the unseen data points (which is also impossible in two-sample testing). During the training of classifiers, we manually assign labels to facilitate distinction by the classifier, whereas in testing, we consider the two samples holistically rather than focusing on individual instance accuracy. \footnotetext{The result does not include standard deviation, since each trial we are testing whether two groups of drawn samples are from same distribution or not, and the result of each trial is either 0 or 1.}

Furthermore, mainstream SSL methods, which primarily enhance classification through data augmentation based on smoothness assumptions or propagate pseudo labels based on clustering assumptions, aim to generate high-confidence training data. However, in two-sample testing, these approaches are flawed; data augmentation may alter the samples' distributions, and pseudo label propagation often proves inaccurate. These discrepancies lead to the frequent ineffectiveness of these SSL methods in two-sample testing contexts. The details of why testing data does not always satisfy the assumptions made by many SSL methods is analyzed in Appendix \ref{not_satisfy_analysis}. Moreover, two-phase representation learning can also be considered as semi-supervised learning \citep{dai2015semisupervisedsequencelearning}, and RL-C2ST is the classifier-based model implemented on the RL-TST framework, so we will also provide more concrete theoretical analysis on how to understand the test power improvement of RL-C2ST in a semi-supervised discriminator's view in Appendix \ref{theory}.

\section{Experiments}


\textbf{Datasets.}
We conducted experiments on five different datasets to thoroughly evaluate our methods in two different aspects: 1) To assess the performance of alternative SSL learning methods directly applied to two-sample testing methods, we utilized three synthetic datasets: \textit{HDGM-Easy}, \textit{HDGM-Medium}, and \textit{HDGM-Hard}. As we have already mentioned in Appendix \ref{discuss_semi}, these datasets represent three different levels of data structure complexity often encountered in the two-sample testing scenarios, which can verify whether the mainstream SSL techniques are robust to various two-sample testing tasks or not; 2) To verify the effectiveness of proposed two-phase RL-TST framework applied on two-sample testing methods (i.e., RL-C2ST, RL-MMD-D) than the other existing work, we conduct the experiments of an implemented RL-TST against other SOTA two-sample testing methods. These experiments were carried out on three representative datasets: \textit{MNIST}, \textit{ImageNet}, and \textit{HDGM-D} (a.k.a. \textit{HDGM-Hard}) to evaluate the enhanced performance of our RL-TST framework. Detailed descriptions of these datasets are provided in Appendix \ref{overview_datasets}.

\begin{figure*}[ht]
    \centering
    \includegraphics[width=1.7\columnwidth]{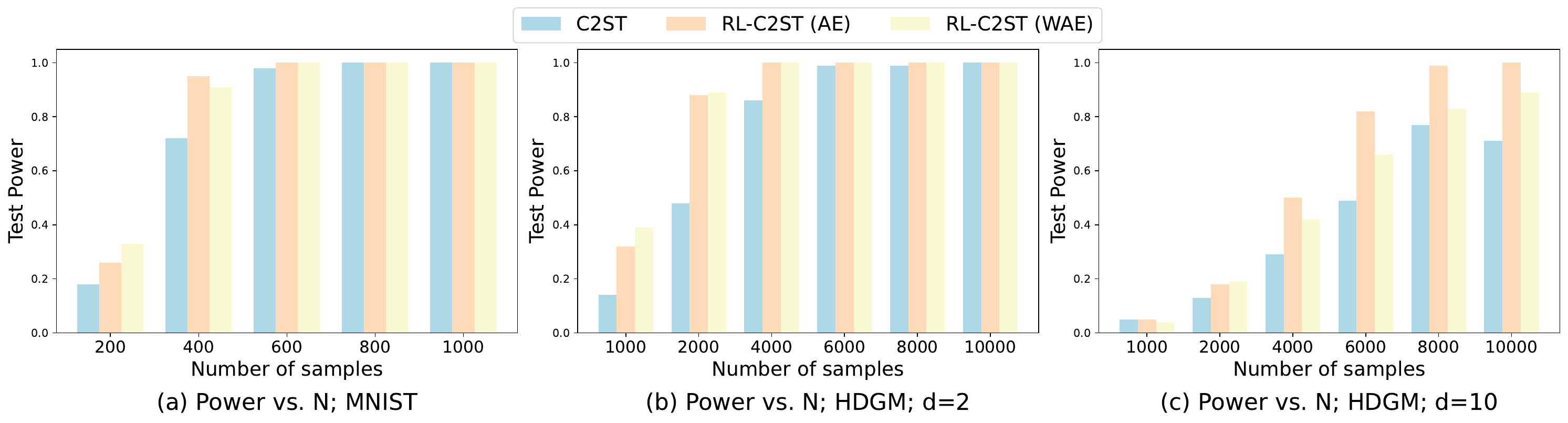}
    \caption{\small Test power of two different implementations of RL-TST framework on the two-sample testing method C2ST. Barplot to show how standard auto-encoder RL-C2ST and wasserstein auto-encoder RL-C2ST both outperform C2ST in the \textit{MNIST} dataset (a), \textit{HDGM-D} when $d=2$ (b) and \textit{HDGM-D} when $d=10$ (c).}
    \label{mnist_barplot}
\end{figure*}

\textbf{Baselines.}
The baselines are the SOTA two-sample testing methods from the existing frameworks. Our main empirical experiments aim to evaluate the performance of two-sample testing methods built on the RL-TST framework (i.e., RL-C2ST and RL-MMD-D) against several SOTA baseline methods in two-sample testing, specifically C2ST, C2ST-L, MMD-D, and MMD-FUSE. These baselines serve as competitive references to highlight the improvements achieved by focusing on learning good representations from RL-TST framework. The following are the overall descriptions of each baseline method,

\begin{itemize}[leftmargin=1.25em,itemsep=3pt,topsep=4pt]
    \item C2ST: C2ST learns a classifier and uses statistic in \eqref{c2st_stats} to measure the difference of two samples \citep{Lopez:C2ST}.
    \item C2ST-L: same as C2ST, except it uses the statistic in \eqref{c2st_l_stats} to measure the absolute mean differences between the probability of the logits of two samples, as we discuss in the Section \ref{preliminary} \citep{cheng2020classificationlogittwosampletesting}.
    \item MMD-D: MMD test trains a neural network to derive a deep kernel \citep{liu2020learning}.
    \item MMD-FUSE: a SOTA testing method that learns IRs from different fixed kernels without data splitting \citep{biggs_mmd-fuse_2023}. 
    \item RL-C2ST: RL-C2ST is a C2ST improved by our proposed RL-TST framework, as we discussed in the Section \ref{proposal}. RL-C2ST-L uses \eqref{c2st_l_stats} as test statistic. 
    \item RL-MMD-D: RL-MMD-D is the implementation of RL-TST on the MMD-D as we discussed in the Section~\ref{proposal}.
\end{itemize}

For their detailed implementations and parameter settings, please refer to Appendix \ref{mmd_based_details}. Moreover, the methods listed in the Table \ref{SOTA-ssl-result} represent our implementations, which serves as an alternative idea's motivation experiment to highlight the challenges of the research gap between the fields of semi-supervised learning and two-sample testing. Thus, they are not intended to be considered as formal baselines.

\textbf{Ablation Study.} 
\label{ablation}
We conduct the ablation study on RL-TST framework by not solely applying one single representation learning algorithm on the original two-sample testing methods. To ensure the effectiveness of the RL-TST framework can be further investigated with other advanced representation learning algorithms,  except for comparing one basic standard AE \citep{Schmidhuber_2015}, we also implement another AE-based representation learning algorithm, wasserstein auto-encoder (WAE) \citep{tolstikhin2019wassersteinautoencoders}, where they both show that various IR learning algorithms can all leverage the information discarded by data splitting to improve the test power of original methods. 

For the details of these two AE-based representation learning, the standard auto-encoder has an unrestricted latent space which can more focus on the reconstruction \citep{Schmidhuber_2015}, while the wasserstein auto-encoder can match the latent space with a target prior distribution (i.e., Gaussian) to make sure the generating power \citep{tolstikhin2019wassersteinautoencoders}. Thus, in application, depending on the characteristics of different AEs, we can choose the suitable one for the downstream task or target data structure. 

For the empirical experiments result, the visualized result of how two kinds of RL-C2STs outperform C2ST is displayed in Figure \ref{mnist_barplot}. In both dataset \textit{MNIST} and \textit{HDGM-Hard}, we can see that the test powers of RL-C2STs are higher than that of C2ST no matter how many numbers of two samples are drawn from the distribution. Although the differences between two methods are little when $N$ is small, the test powers of RL-C2STs have a huge gap over C2ST when $N$ is large enough and converges to 1 with a relative smaller $N$ compare to C2ST. Since under the alternative hypothesis $H_1: \mathbb{P} \neq \mathbb{Q}$, if the number of samples goes to infinity, an effective two-sample testing method will always reject the null hypothesis $H_0: \mathbb{P} = \mathbb{Q}$, thus, the less samples needed to reach the test power of 1, the better performance the method has. Thus, the empirical results can clearly verify that no matter what kinds of representation learning algorithms that can effectively learn IRs from two-sample testing data before learning DRs, we can finally derive a better representation than purely learning DRs. 

\begin{table*}[t]
\centering
\caption{\small MNIST and ImageNet ($\alpha = 0.05$). Average test power for comparing $M$ real MNIST images to $M$ DCGAN-generated MNIST images, and Average test power for comparing $M$ real ImageNet images to $M$ StyleGAN-XL-generated ImageNet images. The three implementations of RL-TST are all using standard auto-encoder in the learning IRs step, we could replace it into other alternative auto-encoders, such as wasserstein auto-encoder discussed in the Section \ref{ablation}.}
\vspace{-0.8em}
\label{image_result}
\resizebox{\textwidth}{!}{%
\begin{tabular}{@{}lllllllllllll@{}}
\toprule
\multicolumn{1}{c}{\multirow{2}{*}{Method}} & \multicolumn{6}{c}{MNIST}                           & \multicolumn{6}{c}{ImageNet}                  \\  \cmidrule(l){2-7} \cmidrule(l){8-13}
\multicolumn{1}{c}{}                        & M=200 & M=400 & M=600 & M=800 & M=1000 & Avg. & M=200 & M=400 & M=600 & M=800 & M=1000 & Avg. \\ \cmidrule(r){1-1} \cmidrule(l){2-7} \cmidrule(l){8-13}
C2ST                                        &      $0.180_{\pm .046}$       &  $0.720_{\pm .023}$     &   $0.980_{\pm .013}$    &    $\textbf{1.000}_{\pm .000}$   &   $\textbf{1.000}_{\pm .000}$     &   $0.776$   &   $0.150_{\pm .022}$    &   $0.300_{\pm .029}$   &   $0.350_{\pm .026}$    &   $0.600_{\pm .036}$    &    $0.850_{\pm .016}$    &   $0.450$   \\
C2ST-L                                      &      $0.250_{\pm .047}$       &   $0.730_{\pm .053}$    &   $0.990_{\pm .009}$    &    $\textbf{1.000}_{\pm .000}$   &    $\textbf{1.000}_{\pm .000}$    &   $0.794$   &   $0.150_{\pm .042}$    &   $0.350_{\pm .030}$    &   $0.450_{\pm .040}$    &     $0.700_{\pm .049}$  &   $0.850_{\pm .034}$     &   $0.500$   \\
MMD-D                                       &      $0.290_{\pm .017}$       &   $0.996_{\pm .009}$    &   $\textbf{1.000}_{\pm .000}$    &   $\textbf{1.000}_{\pm .000}$    &   $\textbf{1.000}_{\pm .000}$     &  $0.857$    &   $0.210_{\pm .031}$    &  $0.400_{\pm .039}$     &    $0.570_{\pm .033}$   &   $0.780_{\pm .041}$    &   $\textbf{1.000}_{\pm .000}$     &  $0.592$    \\
MMD-FUSE                                    &     $0.320_{\pm .032}$        &   $0.870_{\pm .033}$    &  $\textbf{1.000}_{\pm .000}$     &  $\textbf{1.000}_{\pm .000}$     &   $\textbf{1.000}_{\pm .000}$     &   $0.838$   &   $0.230_{\pm .029}$    &   $0.450_{\pm .034}$    &   $0.610_{\pm .037}$    &   $0.790_{\pm .029}$    &   $\textbf{1.000}_{\pm .000}$     &   $0.616$   \\ \midrule
RL-C2ST                                    &     $0.260_{\pm .049}$        &   $0.950_{\pm .022}$    &   $\textbf{1.000}_{\pm .000}$    &  $\textbf{1.000}_{\pm .000}$     &    $\textbf{1.000}_{\pm .000}$    &   $0.842$   &   $0.200_{\pm .036}$    &   $0.400_{\pm .049}$    &    $0.500_{\pm .061}$   &   $0.650_{\pm .050}$    &    $0.950_{\pm .022}$    &  $0.540$    \\
RL-C2ST-L                                  &     $\textbf{0.491}_{\pm .060}$        &  $0.985_{\pm .013}$     &   $\textbf{1.000}_{\pm .000}$    &   $\textbf{1.000}_{\pm .000}$    &   $\textbf{1.000}_{\pm .000}$     &   $\textbf{0.895}$   &   $\textbf{0.400}_{\pm .059}$    &   $\textbf{0.500}_{\pm .059}$    &   $0.650_{\pm .056}$    &   $0.750_{\pm .054}$    &    $\textbf{1.000}_{\pm .000}$    &   $0.660$   \\ 
RL-MMD-D                                  &     $0.420_{\pm .072}$        &  $\textbf{1.000}_{\pm .000}$     &   $\textbf{1.000}_{\pm .000}$    &   $\textbf{1.000}_{\pm .000}$    &   $\textbf{1.000}_{\pm .000}$     &   $0.884$   &   $0.330_{\pm .051}$    &   $0.470_{\pm .069}$    &   $\textbf{0.680}_{\pm .055}$    &   $\textbf{0.890}_{\pm .037}$    &    $\textbf{1.000}_{\pm .000}$    &   $\textbf{0.674}$   \\ 
\bottomrule
\end{tabular}%
}
\end{table*}

In an analytical view, compared to C2ST, both RL-C2STs learn a compact and potentially more informative representation of the whole data, which makes efficient use of the unlabelled test data. This can not only discover underlying patterns or features that might not directly related to the labels but to the data distribution itself, but also provide a regularizing effect to prevent the model being more likely to overfit the training data. Similar to all of the applications of RL-TST, such featurizer in the RL-TST can result in a better generalization from the learned representations and improve the classifier's performance on the testing set predictions. The empirical outperformance of different learning algorithms also validates that RL-TST is a compelling framework for two-sample testing methods learning from unlabelled data.

\begin{figure}[t]
\begin{center}
\centerline{\includegraphics[width=0.9\columnwidth]{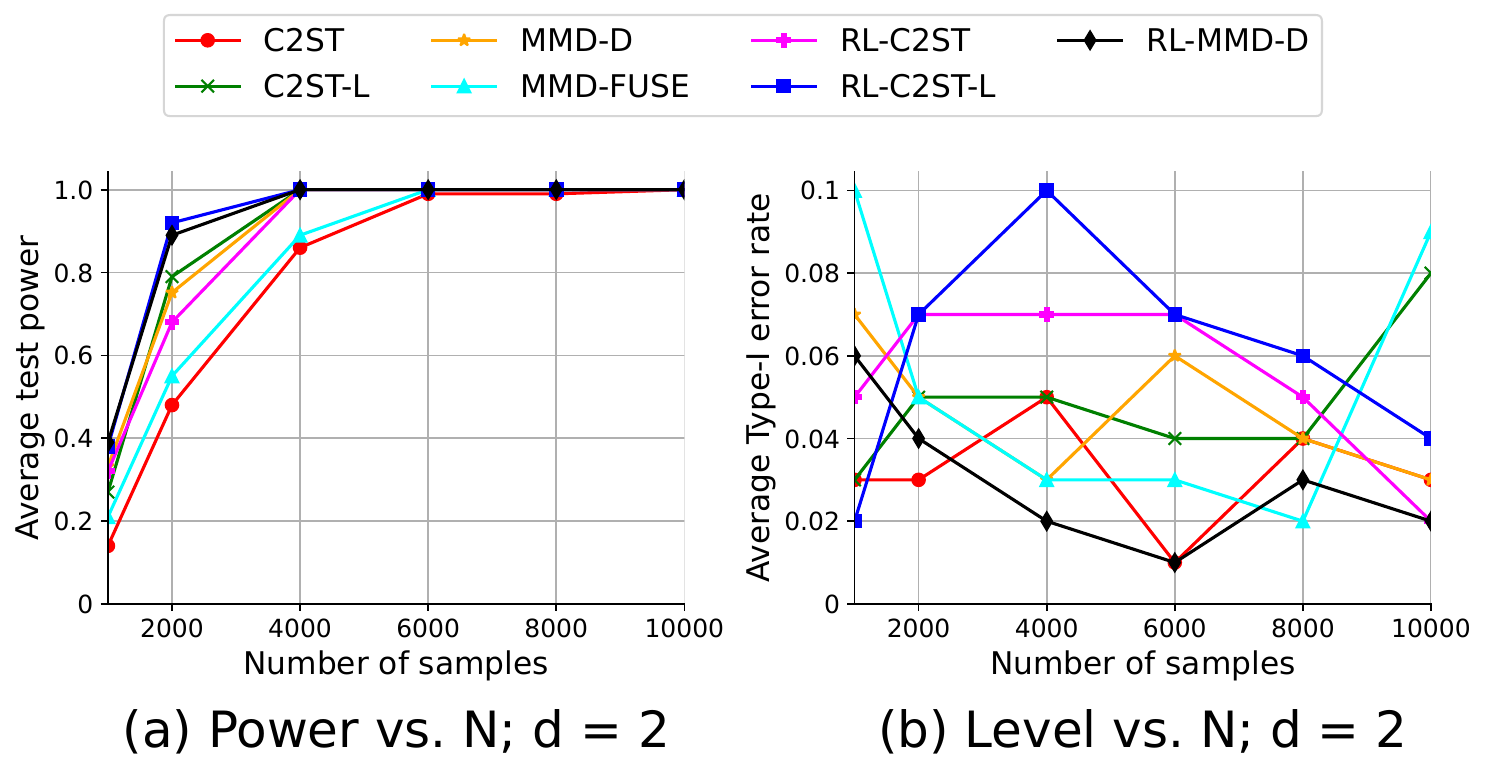}}
\caption{Results on \textit{HDGM-D} and \textit{HDGM-S} for $\alpha = 0.05$. (a) average test power and (b) average type-I error keeping $d=2$ in 100 trials when increasing $N$ from $N=1000$ to $N=10000$. The RL-TST methods are all using the standard auto-encoder, we could replace it into other alternative auto-encoders, such as wasserstein auto-encoder as we discussed in the Section \ref{ablation}.} 
\label{HDGM_result}
\end{center}
\end{figure}

\textbf{Result Analysis.}
After we validate the effectiveness of RL-TST on C2ST, we will also display how effective the RL-TST applied on two advanced two-sample testing methods, which results in RL-C2ST-L and RL-MMD-D. They both show not only how they improve the original C2ST-L and MMD-D, but also how they outperform the most SOTA testing method MMD-FUSE. 

The overall result of all testing methods for the HDGM dataset is shown in Figure \ref{HDGM_result}. We can see that all the RL-TST methods have the higher test power than the original method, no matter how we choose $N$, while all type-I errors are reasonably controlled around $\alpha=0.05$. For \textit{MNIST} and \textit{ImageNet} datasets, the results of all methods are shown in Table \ref{image_result}, all the RL-TST methods still outperform the original methods. Moreover, RL-C2ST-L and RL-MMD-D, which are the most two powerful applications, can have the highest test power than other SOTA methods among different sample size $N$ or $M$, which can verify the improvement of our RL-TST on different two-sample testing methods. 

\textbf{Discussion of Sequential Two-sample Testing.}
Sequential two-sample testing methods also utilize information from the test data but has a different problem setting from what we are interested in. In our problem setting, we assume the total number of samples is \emph{fixed and given}, and we are trying to distinguish whether these two given samples are from the same distribution or not. No more extra data are provided for testing data and the test data is known, so it can be regarded as a \emph{transductive learning problem}, while the sequential two-sample testing assumes the testing data can infinitely arrive as batch. We provide detailed descriptions of sequential two-sample testing in Appendix \ref{sequential_tst}, along with experimental results in Appendix \ref{sequential_result} demonstrating that RL-TST can outperform these sequential approaches within the same setting.


\section{Conclusion}
Non-parametric two-sample testing is an important problem in both statistics and machine learning fields. This paper presents a unified view, \emph{focusing on learning good representations from both labelled and unlabelled samples}, to both leverage the discarded information in the data splitting process and enhance the discriminative ability, which can address the existing drawbacks of two-sample testing methods. In order to examine the viability of the view, we conduct a thorough survey in the field of two-sample testing and the potential fields that enable to utilize information from unlabelled data, and propose a feasible framework that empirically improve the performance of two-sample testing methods. In the future, more advanced representation learning techniques for two-sample testing methods can be developed based on this proposed research direction.

\bibliography{uai2025-template}

\newpage

\onecolumn


\appendix

\section{Algorithm} \label{algorithm}
We present the general framework for RL-TST in the following Algorithm \ref{rl-tst-algorithm}.
\begin{algorithm}[]
   \caption{Paradigm of testing with RL-TST}
   \label{rl-tst-algorithm}
\begin{algorithmic}
    \STATE {\bfseries Input:} $S_\mathbb{P}$, $S_\mathbb{Q}$, significance level $\alpha$, an auto-encoder $f_{a}$ consist of a featurizer $\phi$ and a decoder $\phi^{-1}$, a final classifier $\mathcal{M} := g \circ \phi$ or a deep kernel $\mathcal{M} := k_{\phi}$ included with the featurizer, total epochs for learning IRs $T_{\rm IR}$, total epochs for learning DRs $T_{\rm DR}$. \\
    \STATE \textbf{1:} Derive the unlabelled data $\mathcal{S}_{\rm unl} = \text{shuffle} (S_\mathbb{P} \cup S_\mathbb{Q})$ \\
    \leftcomment{Phase 1: derive Featurizer $\phi$ from learning IRs}\\
    \FOR{$t = 1, 2, \dots, T_{\rm IR}$} 
        \STATE \textbf{2:} $X_t \gets \text{minibatch from } \mathcal{S}_{\rm unl};$
        \STATE \textbf{3:} $\phi^* \gets \arg\min_\phi \mathcal{R}(f_a, X)$ based on \eqref{IR};
    \ENDFOR \\
    \leftcomment{Phase 2: train a classifier or kernel $\mathcal{M}$ to learn DRs on $S^{\rm tr} = (S_\mathbb{P}^{\rm tr}, \mathbf{0}) \cup (S_\mathbb{Q}^{\rm tr}, \mathbf{1})$} \\
    \FOR{$t = 1, 2, \dots, T_{\rm DR}$} 
        \STATE \textbf{5:} $(X_t, l_t) \gets \text{minibatch from } \mathcal{S}^{\rm tr};$
        \STATE \textbf{6:} $g^* \gets \arg\min_g \mathcal{L}_{\rm DR}(\phi^*(X_t), l_t, g)$ based on ~\eqref{DR_clf}, if learning classifier; 
    \ENDFOR \\
    or \STATE \textbf{7:} $k^* \gets \arg\max_{k_{\phi^*}} \widehat{\mathcal{J}}_{\rm DR}(S_\mathbb{P}^{\rm tr}, S_\mathbb{Q}^{\rm tr}; k_{\phi^*}) $ based on \ref{DR_mmd}, if learning deep kernel; \\
    \leftcomment{Phase 3: permutation test with $f$ on $S^{\rm te} = S_\mathbb{P}^{\rm te} \cup S_\mathbb{Q}^{\rm te}$} \\
    \STATE \textbf{8:} $est \gets \hat{t}(S_\mathbb{P}^{\rm te}, S_\mathbb{Q}^{\rm te}; \mathcal{M})$ based on~\eqref{c2st_stats}, \eqref{c2st_l_stats}, or \eqref{mmd_stats};
    \FOR{$i = 1, 2, \dots, n_{\rm perm}$} 
        \STATE \textbf{9:} Shuffle $S^{\rm te}$ into $X$ and $Y;$ \\
        \STATE \textbf{10:} $perm_i \gets \hat{t}(X, Y; \mathcal{M})$ 
    \ENDFOR
    \STATE {\bfseries Output:} $\mathbb{I}\left[\frac{1}{n_{perm}}\sum_{i=1}^{n_{perm}}\mathbb{I}(est < perm_i) \leq \alpha\right]$
\end{algorithmic}
\end{algorithm}

\section{Discussion of Semi-supervised Learning}
\label{discuss_semi}

\subsection{Overview of Main Categories of Semi-supervised Learning Methods}
\label{overview_sota_ssl}
Building on the semi-supervised learning (SSL) assumptions, we will recap how contemporary SOTA SSL methods incorporate these principles and assumptions, setting the stage for an analysis of their applicability to the specific challenges presented by our problem setting.

\textbf{Transductive vs Inductive learning.}  Classification tasks within machine learning can typically be categorized within two distinct problem settings: transductive and inductive learning \citep{mit06ssl}. \emph{Transductive learning} is concerned with predicting the labels of the specific unlabelled data that was present during the training process, emphasizing a tailored fit to this data. \emph{Inductive learning}, on the other hand, focuses on the generalization of the learned classifier to new, unseen data. In learnable two-sample testing, the goal is to test whether the given two samples are drawn from same distributions. To make it, we firstly split samples into labelled set and unlabelled set, then find out that whether it is possible to learn a classifier that can distinguish two samples from the mixed unlabelled set. It becomes apparent that applying SSL methodologies to the two-sample testing problem inherently requires a transductive learning approach. This conceptual groundwork necessitates a detailed examination of current SSL methods to identify their foundational assumptions and evaluate their performance in two-sample test scenarios. 

\textbf{Major categories.} Currently, we identify that there are five main categories of SOTA SSL methods: consistency regularisation, pseudo-labelling, graph-based, generative models and hybrid (often a combination of consistency regularisation and pseudo-labelling) \citep{yang2023sslsurvey}. We will succinctly explicate how they work, and how they are applied for our downstream two-sample testing tasks in the experiments of various levels of HDGM.
\begin{itemize}
    \item \textit{Consistency Regularisation:} Based on the manifold assumption or the smoothness assumption, the consistency regularisation methods apply consistency constraints to the final loss function, where the intuition is that if the data follows the smoothness assumption or manifold assumption, even though we construct some perturbations in the inputs, it will not influence the output of classification \citep{xie2020unsupervised}.
    
    \item \textit{Pseudo-Labelling:} Pseudo-labelling uses its own predictions to generate labels for unlabelled data, which are then used to further train the model. It relies on the assumptions that model's high-confidence predictions are accurate. This assumption is based on the cluster assumption for the validity and efficacy of propagating labels to unlabelled data based on model predictions \citep{lee2013pseudo}.

    \item \textit{Graph-Based:} Graph-based methods will construct a similarity graph based on the raw dataset, where each node represents a data instance, and weighted-edge represents the similarity between two data instances. Based on the smoothness assumption, the label information can be propagated from labelled nodes to unlabelled nodes, if two nodes are closely connected in the constructed graph \citep{song2021graphbased}.

    \item \textit{Generative Models:} Generative methods learn to model the underlying distribution of both labelled data and unlabelled data, using this learned representation to generate new data points and infer missing labels. Based on the manifold assumption, the generative models aim to learn the underlying low-dimensional manifold and generate data points that adhere to the same manifold, used for further model training \citep{kingma2013auto}.

    \item \textit{Hybrid:} Hybrid methods are just combination of multiple methods, such as consistency regularisation, pseudo-labelling, and sometimes generative approaches. These models typically rely on the smoothness assumption and cluster assumption, in order to infer the labels of unlabelled data \citep{sohn2020fixmatch}.
    
\end{itemize}

\subsection{Analysis of Why Two-sample Testing Data are Not Satisfied for SSL}
\label{not_satisfy_analysis}


In the traditional two-sample testing problem settings, there is often overlap between the two samples. As we can see in Figure~\ref{HDGM_overview}b and Figure~\ref{HDGM_overview}c, for the HDGM-Medium and HDGM-Hard datasets, there are high-overlapping areas between two distributions. This will \textit{highly violate the first two assumptions} of SSL mentioned previously. For the smoothness assumption, our dataset will have large amounts of nearly the same data points in two samples, but allocated different labels; this will notably influence the SSL methods that based on such assumption. For the cluster assumptions, we can see in HDGM-Medium that although there are two obvious clusters, they do not have the same labels within the same cluster in a holistic view. The SOTA SSL techniques will rely on at least one of the smoothness assumption or cluster assumption to ensure that the unlabeled samples' label information can be inferred, or extra training samples can be created. However, our samples will face a challenge that they may only \emph{follow the manifold assumption}: to ensure the robustness of the methods, we have to make sure that they can be applied on all the possible scenarios.

\section{Experimental Details} \label{experiment_details}

\subsection{Overview of Datasets} \label{overview_datasets}
\textbf{High-Dimensional Gaussian mixtures.} 
The \emph{high dimensional Gaussian mixtures} (HDGM) benchmark is a synthetic dataset that is composed of multiple Gaussian distributions, each representing a cluster, which is proposed by \citet{liu2020learning}. In our experiments, we are considering bimodal Gaussian mixtures, which means the number of clusters remains 2 irrelevant to the dimension of the multivariate Gaussian distributions. In Appendix \ref{discuss_semi}, we consider there are three levels of \textit{HDGM}, which are \textit{HDGM-Easy}, \textit{HDGM-Medium} and \textit{HDGM-Hard} in order to specify different levels of data distribution existing in the two-sample testing problems. In other places of this paper, rather than Appendix \ref{discuss_semi}, we regard \textit{HDGM} as \textit{HDGM-Hard}. Under $H_0$, $\mathbb{P}$ and $\mathbb{Q}$ are the same, which denoted as \textit{HDGM-S} to verify the type I error under control; and under $H_1$, we slightly modify a mild covariance $\pm0.5$ between first two dimensions in the covariance matrix of $\mathbb{Q}$ and other setups are the same as \textit{HDGM-S}, which is referred to as \textit{HDGM-D}. Thus, \textit{HDGM-S} and \textit{HDGM-D} are both noted by hard-level HDGM. The details of how to synthesize $\mathbb{P}$ and $\mathbb{Q}$ to derive \textit{HDGM-Easy}, \textit{HDGM-Medium}, \textit{HDGM-Hard}, \textit{HDGM-S} and \textit{HDGM-D} are described in Appendix \ref{hdgm_dataset_detail}. We regard $n_c$ as the number of samples drawn from each cluster in each distribution and $N$ as the number of total samples drawn from both $\mathbb{P}$ and $\mathbb{Q}$, where $N = n \times c \times 2$. We conduct two experiments on $\textit{HDGM-D}$, increasing the $N$ from $N = 1000$ to $N = 10000$ when keeping the dimension $d$ remain the same. One experiment is a low-dimensional $\textit{HDGM-D}$ with $d=2$ and another is a high-dimensional $\textit{HDGM-D}$ with $d=10$. Moreover, we conduct both low-dimensional and high-dimensional $\textit{HDGM-S}$ to show that the type-I error is controlled. The result is shown in Figures \ref{mnist_barplot} and \ref{HDGM_result}, which will be analyzed in the below subsection.

\textbf{\textit{MNIST} vs \textit{MNIST}-Fake.} The \textit{MNIST} datasets is a collection of 70,000 grayscale images of handwritten digits, ranging from 0 to 9, divided into a training set of 60,000 images and a test of 10,000 images \citep{lecun1998gradient}. The \textit{MNIST}-Fake is the a set of 10,000 images generated by a pretrained \emph{deep convolutional generative adversarial network} (DCGAN) \citep{DCGAN_Radford}. The MNIST benchmark (\textit{MNIST} vs \textit{MNIST}-Fake) is also proposed by \cite{liu2020learning}, aiming to test the performance of testing methods in the image space. Under $H_0$, we draw samples both from the \textit{MNIST}-Fake. Under $H_1$, we compare the samples from real \textit{MNIST}, $\mathbb{P}$, and samples from \textit{MNIST}-Fake, $\mathbb{Q}$. We regard $N$ as the number of samples each drawn from $\mathbb{P}$ and $\mathbb{Q}$, where we increase $N$ from $N=200$ to $N=1000$. The result of the average test power of all methods is displayed in the Table \ref{image_result}.  All methods are tested with a reasonable type-I error rate.

\textbf{ImageNet vs ImageNet-Fake.} The \textit{ImageNet} dataset is a comprehensive collection of over 14 million labelled high-resolution images belonging to roughly 22,000 categories \citep{deng2009imagenet}. The \textit{ImageNet}-Fake dataset comprises 10,000 high-quality images generated using the advanced \emph{StyleGAN-XL} model, a state-of-the-art generative adversarial network designed for large and diverse datasets \citep{sauer2022stylegan}. This benchmark (\textit{ImageNet} vs \textit{ImageNet}-Fake) extends the framework established by \citet{liu2020learning} to a more complex and diverse image domain, testing the robustness of two-sample testing methods at a larger scale. Under the null hypothesis $H_0$, samples are drawn from \textit{ImageNet}-Fake, while under the alternative hypothesis $H_1$, we compare samples from the real \textit{ImageNet} dataset, $\mathbb{P}$, with those from \textit{ImageNet}-Fake, $\mathbb{Q}$. We vary the number of samples drawn from each, $\mathbb{P}$ and $\mathbb{Q}$, from $N = 200$ to $N = 1000$ to examine the scalability of the test methods. The outcomes in terms of average test power across various methodologies are summarized in Table \ref{image_result}, with all tests maintaining a reasonable type-I error rate.

\subsection{Implementation Details of C2ST and RL-C2ST} \label{ablation_details}
\begin{itemize}
    \item C2ST: C2ST has the ability of learn DRs from two samples, by learning a well-trained classifier to only get prediction accuracy information. Implementation of C2ST paradigm is to only take \textit{Phase 2} and \textit{Phase 3} from Algorithm \ref{rl-tst-algorithm}. Most of the implementation details are referenced from \cite{Lopez:C2ST} and \cite{liu2020learning}. The splitting portion of training and testing is always half to half, and the model architecture is the same for C2ST and RL-C2ST, where first few layers are feature extractor and followed by a classification layer. Moreover, in the first step of \textit{Phase 3}, we do not utilize the the softmax probability of the first value of the logits returned by the classifier to calculate the statistic of two samples, we apply \eqref{c2st_stats} which directly derive the mean of the classification prediction accuracy of two samples.
    \item RL-C2ST: a unified representation learning version of C2ST. We firstly learn an encoder that can extract IRs from whole samples, and boost the discriminative ability by minizing the prediction error of classifier. Replacing the test statistics of RL-C2ST from \eqref{c2st_stats} into \eqref{c2st_l_stats} will result in RL-C2ST-L. Most of the implementation details are described in the Algorithm \ref{rl-tst-algorithm}.
\end{itemize}
In C2ST, we have a classifier $f$ consisting of a randomly initialized feature extractor $\phi_{\theta}(x)$ followed by a logistic regression layer with parameters $\bm{w}$ and $\bm{b}$, where
\[ 
    f(x) = \phi_{\theta}(x) \times \bm{w} + \bm{b}. 
\]
As the $f$ is a binary classifier, $f(x) = [z_0, z_1]$ and $\text{softmax}(f(x)) = [p_0, p_1]$, where $p_0 + p_1 = 1$. All parameters $\theta, \bm{w}$ and $\bm{b}$ are updated through the supervised learning on the training set, which aims to minimize the occurrence of incorrect predictions.
Then, use the empirical probability of the correct predictions on an unseen testing set to measure the difference between two samples.

However, in RL-C2ST, we have $g$ consisting of a feature extractor $\phi_a(x)$ trained on $S_P^{\rm tr} \cup S_P^{\rm te} \cup S_Q^{\rm tr} \cup S_Q^{\rm te}$ without labels via unsupervised learning and a logistic regression layer for subsequent supervised training purpose. In the unsupervised learning step, we use $\phi_{a}(x)$ to extract a latent feature vector $z$ from the input $x$, and then use a decoder $\phi^{-1}_a(x)$ to reconstruct $z$ to a reconstructed $x'$. We update the parameters of $\phi_a$ by minimizing the difference between the reconstructed input $x'$ and the original input $x$. After the unsupervised training procedure, we add a classification layer after $\phi_{a}$ to form a classifier $g$, and train the classification layer in the same way as the C2ST.

\subsection{Details of RL-C2ST-L and other MMD based methods} \label{mmd_based_details}
We first introduce RL-C2ST-L and compare the following state-of-the-art testing methods on two benchmark datasets:
\begin{itemize}
    \item C2ST-L: The name of C2ST-L is originated from \cite{liu2020learning}, where L refers to logit. It can capture more discriminative information from the confidence of predictions. The implementation detail is the same as C2ST, except not computing the prediction probability, but using the logit output directly to measure the distance. 
    \item RL-C2ST-L: A RL-TST implemented on C2ST-L. Rather than using the prediction labels (0 or 1) to measure the test accuracy, we utilize MMD to calculate the differences between output features extracted from the RL-C2ST. The output features could be the output of the hidden layer or the logits output of the classifier trained by the RL-C2ST, as we discuss in Section \ref{proposal}.
    \item MMD-D: a SOTA testing method to learn a deep kernel with a neural network that can extract DRs. MMD-D has the training objectives of directly maximizing the test power of MMD, leading to an increase in test power on the testing set. The implementation is strictly aligned with the code provided in \cite{liu2020learning}, where we simultaneously train a deep neural network and deep kernel Gaussian bandwidths by maximizing the training objectives $\widehat{\mathcal{J}} = \widehat{\rm MMD}_u / \hat{\sigma}_{\mathcal{H}_1, \lambda}$. 
    \item RL-MMD-D: A RL-TST implemented on MMD-D to alleviate the drawbacks of decreasing the testing samples size from data splitting process, as we discussed in the Section \ref{proposal}.
    \item MMD-FUSE: MMD-FUSE fuses several MMD statistics based on the simple kernel of different combinations of hyperparameters into a new powerful statistic, then conducts a permutation test based on the fused statistic. The implementation is strictly aligned with the code provided in \cite{biggs_mmd-fuse_2023}, where we compute and fuse the test statistics based on different kernel functions and hyperparameters in order to capture complex data structure in an unsupervised way.
\end{itemize}

\subsection{Details of HDGM datasets} \label{hdgm_dataset_detail}
Table \ref{HDGM_details} displays the details of how HDGM datasets are generated \citep{liu2020learning}. Different levels of HDGM datasets are first proposed in this paper, in order to show why SOTA SSL methods cannot be directly applied in the two-sample testing problem. The level of HDGM is differed from whether the data points are highly overlapping or whether the clusters within the same distribution are isolated. For the \textit{HDGM-Easy}, $\Delta_{\mu} = 10$ and $\Delta_q = 5$. For the \textit{HDGM-Medium}, $\Delta_{\mu} = 10$ and $\Delta_q = 0$. For the \textit{HDGM-Hard}, $\Delta_{\mu} = 0.5$ and $\Delta_q = 0$.

\begin{table}[ht]
    \vspace{-1em}
    \centering
    \caption{Details of how to synthesize $\mathbb{P}$ and $\mathbb{Q}$ in the experiments. Let $c = 2$ be the number of the clusters in each distribution, $d > 2$ be the dimension of multivariate normal distribution of each cluster. $\left(\bm{\mu}_1, \dots, \bm{\mu}_c\right)$ is a set of d-dimensional mean vector $\bm{\mu}_i$ that specifies that mean of each dimension in the distribution, where $\bm{\mu}_{1} = \bm{0}_d, \bm{\mu}_{i} = \bm{\mu}_{i-1} + \Delta_{\mu} \times \bm{1}_d$. $I_d$ is the $d \times d$ identity matrix, $\Delta_{\mu}$ is the cluster mean difference within the same distribution, and $\Delta_{q}$ is the mean difference between $\mathbb{P}$ and $\mathbb{Q}$. $\Delta_1 = 0.5$, $\Delta_2 = -0.5$, and $\bm{\Sigma}_i = 
        \begin{pmatrix}
            1 & \Delta_i & \bm{0}_{d-2}\\
            \Delta_i & 1 & \bm{0}_{d-2}\\
            \bm{0}_{d-2}^T & \bm{0}_{d-2}^T & I_{d-2}
        \end{pmatrix}$.}
    
    \label{HDGM_details}
    
    \begin{tabular}{l|ll}
        \toprule
        Datasets & $\mathbb{P}$ & $\mathbb{Q}$ \\
        \midrule
        \textit{HDGM-S} & $\sum_{i=1}^c \mathcal{N}\left(\bm{\mu}_i, I_d\right)$ & $\sum_{i=1}^c \mathcal{N}\left(\bm{\mu}_i, I_d\right)$  \\
        \textit{HDGM-D} & $\sum_{i=1}^c \mathcal{N}\left(\bm{\mu}_i, I_d\right)$ &  $\sum_{i=1}^c \mathcal{N}\left(\bm{\mu}_i+\Delta_{q}, \bm{\Sigma}_i \right)$ \\

        \bottomrule
    \end{tabular}
    \vspace{-1em}
\end{table}

\subsection{Details of computing resources} \label{resource}
The experiments of the work are conducted on three platforms. One platform is a Nvidia-4090 GPU PC with Pytorch framework. The second platform is a High-performance Computer cluster with lots of Nvidia-A100 GPU with Pytorch framework. The last platform is a Nvidia-4090 GPU Window Subsystem for Linux with Jax framework. The memory of three platforms are all over 16 GB. The storage of disk of three platforms are all over 512 GB. 

\subsection{A discussion about supervised sequential two-sample testing and applicability of RL-TST}
\label{sequential_tst}
Supervised sequential two-sample testing represents another approach to utilizing testing data \citep{pandeva2024evaluatingclassifiertwosampletests}. In this framework, a classifier is trained to determine whether two samples from a single batch originate from the same distribution. Initially, batches are split and fed sequentially into the classifier as testing data. Batches that do not reject the null hypothesis are concatenated with previous batches and used as training data for the classifier, continuing until all batches are exhausted or a single batch rejects the null hypothesis. The sequential nature of the test emerges from the use of e-values, which are updated as more data becomes available, allowing for a dynamic assessment of the testing hypothesis. However, this method should not be directly compared to our method due to different problem settings and designs. Firstly, in sequential two-sample testing, data are split into several batches and tests are conducted on single, small batches. Conversely, in other supervised two-sample testing approaches, data are only split into two halves, creating a trade-off between the number of training and testing samples.

Furthermore, the design of our RL-TST framework is compatible with any other supervised two-sample testing framework, including sequential two-sample testing. As long as a proportion of data is used for testing, we can remove the labels from this testing data and concatenate it into the training data. This allows us to learn IRs through representation learning, followed by the original supervised two-sample testing framework.

\subsection{Experiment Result of Sequential Two-sample Testing}
In this part, we will display the result of supervised sequential two-sample test that proposed by \citet{pandeva2024evaluatingclassifiertwosampletests} on the HDGM-Hard dataset, and compared the result with original C2ST and RL-C2ST in our problem setting. We can find that even though this method can have a small increase on the test power over the original C2ST method, but have a large decrease to our method. The number of batches we choose is five, if we choose the number of batches to two, it is exactly similar as C2ST; if we choose the number of batches to a large number like ten, the test power will drop down, since the test data size will be too small. Thus, we decide five as the number of batches, and C2ST-Sequential(5) in the Table~\ref{sequential_experiments} represent the supervised sequential two-sample testing with the number of batches equal to five.
\label{sequential_result}
\begin{table}[h]
\vspace{-0em}
\centering
\caption{Experiment results of test power of sequential two-sample testing with Batch5 over original C2ST and our propose RL-C2ST on HDGM-hard dataset. $N$ is the total size of two samples inputed in 100 trials.}
\label{sequential_experiments}
\begin{tabular}{lllll}
\toprule
Method                 & N=4000 & N=6000 & N=8000 & Avg. \\ \midrule
C2ST-Sequential (5) & 0.32   & 0.57   & 0.79   & 0.56 \\
C2ST                   & 0.29   & 0.49   & 0.78   & 0.52 \\
RL-C2ST               & 0.50   & 0.81   & 0.99   & 0.77 \\ \bottomrule
\end{tabular}
\end{table}


\subsection{Reproducibility}
All the reproducible code can be found in the anonymous \href{https://anonymous.4open.science/r/Revisit-non-parametric-two-sample-testing-as-a-semi-supervised-learning-problem-6D0D/README.md}{link}, and some of the two-sample testing methods are used in the package AdapTesting.

\section{Theoretical Discussion and Analysis} \label{theory}
The following theoretical discussions are based on the assumption that the input two-sample testing data can follow the assumptions of the applied semi-supervised methods. Moreover, those theorems are only applied to MLP-based two-sample testing methods, such as RL-C2ST or RL-C2ST-L.
\subsection{Theoretical Declaration and Interpretation} \label{theory_declare}

\textbf{Test Power.} Test power is the probability that a test will correctly reject $H_0$, when $H_1$ holds. It represents the ability of the test to detect the difference between $\mathbb{P}$ and $\mathbb{Q}$, so analyzing this power is essential for evaluating the performance of one two-sample testing method. 
\begin{definition}
    Let $f' \in \mathcal{C}_{\phi} : \mathcal{X} \rightarrow \{0, 1\}$ denotes the RL-C2ST classifier model with specific feature extractor $\phi$, where $\mathcal{C}_{\phi} = \{f' | f' = g \circ \phi, g \in \mathcal{G}\} \subseteq \mathcal{C}$ and $\mathcal{C} = \bigcup_{\phi \in \mathcal{F}} \mathcal{C}_\phi$.
\end{definition}

\begin{theorem}\citep{Lopez:C2ST} 
\label{c2st_test_power}
     Let $H_0: t = \frac{1}{2}$ and $H_1: t = 1 - \epsilon(\mathbb{P}, \mathbb{Q}; f')$, where $t$ is the test accuracy and $\epsilon(\mathbb{P}, \mathbb{Q}; f') = {\rm Pr}_{(z_i, l_i) \sim \mathcal{D}} \left[f'(z_i) \neq l_i \right] / 2 \in \left(0, \frac{1}{2}\right)$ represents the inability of $f'$ to distinguish between $\mathbb{P}$ and $\mathbb{Q}$. The test power of $\hat{t}$  is:
    \begin{equation}
    {\rm Pr}_{H_1}\left(\hat{t}_{H_0} > t_{\alpha}\right) = \Phi \left(\frac{\left(\frac{1}{2} - \epsilon(\mathbb{P}, \mathbb{Q}; f')\right)\sqrt{n_{\rm te}} - \Phi^{-1}(1-\alpha) / 2}{\sqrt{\epsilon(\mathbb{P}, \mathbb{Q}; f') - \epsilon(\mathbb{P}, \mathbb{Q}; f')^2}}\right),
    \end{equation}
    where $\alpha \in (0, 1)$ is the significance level, $t_{\alpha}$ is the $(1-\alpha)$ quantile and $\Phi$ is the CDF of standard normal distribution. The Type-I error of $\hat{t}$ is also controlled no more than $\alpha$, which ensures that the test will not always reject $H_0$, when $H_0$ is true.
\end{theorem}

\textbf{Understand RL-C2ST via Theorem~\ref{c2st_test_power}.} In hypothesis testing, our primary aim is to maximize test power while maintaining control over the Type-I error rate. While we know that via Theorem \ref{c2st_test_power}, $\Phi^{-1}(1-\alpha)/2$ is a constant, for a reasonably fixed large $n_{\rm te}$, the first term $(\frac{1}{2} - \epsilon(\mathbb{P}, \mathbb{Q}; f'))$ in the numerator dominates the test power. In fact, to ensure that the model can achieve the optimal test power on a fixed test dataset, it is equivalent to minimize
\begin{equation}
    \label{objectives}
    \mathcal{J}\left(\mathbb{P}, \mathbb{Q}; f'\right) := {\epsilon(\mathbb{P}, \mathbb{Q}; f')}\Big/{\left(1 - \epsilon(\mathbb{P}, \mathbb{Q}; f')\right)}, 
\end{equation} 
where we estimate it with
\begin{equation}
    \label{approximate_test_power}
    \hat{\mathcal{J}}\left(S_P, S_Q; f'\right) := \frac{\hat{\epsilon}(S_P, S_Q; f')}{\left(1 - \hat{\epsilon}(S_P, S_Q; f')\right)},~\textnormal{and}~\hat{\epsilon}(S_P, S_Q; f') \in  \left(0, \frac{1}{2}\right),
\end{equation}

where $$\hat{\epsilon}(S_P, S_Q; f') = \frac{1}{2}\widehat{err}(f') = \frac{1}{2|\mathcal{S}|}\sum_{(x_i, l_i) \sim \mathcal{S}}\mathbb{I}[f'(x_i) \neq l_i].$$ From \eqref{approximate_test_power}, we can find that if we can learn a classifier $f'$ from \eqref{DR_clf} that has a smaller $\hat{\epsilon}(S_P, S_Q; f')$, we can minimize the $\hat{\mathcal{J}}$, leading to maxmizing the test power. Thus, we will analyze how the use of unlabelled data and the size of unlabelled data $m_{\rm u}$ helps to learn a classifier model $f'$ that have a smaller $\hat{\epsilon}(S_P, S_Q; f')$ in the semi-supervised learning. 

We first give a definition of compatibility, an important measurement when analyzing SSL methods.

\begin{definition}[Compatibility]
    The compatibility of classifier model $f$ is defined as $\chi: \mathcal{C} \times \mathcal{X} \rightarrow [0, 1]$, and $\chi(f, \mathcal{D}) = \mathbb{E}_{x \sim \mathcal{D}}[\chi(f, x)]$ estimates how ``compatible" the $f$ is with $\mathcal{D}$. Thus, for a given sample $\mathcal{S}$, the incompatibility of $f$ with $\mathcal{S}$ is $1 - \chi(f, \mathcal{S})$. We can also call it unlabelled error rate $err_{\rm unl}(f)$, where $\widehat{err}_{\rm unl}(f) = 1 - \chi(f, \mathcal{S})$, e.g., for the consistency regularization technique, $1 - \chi(f, x) = (f(x) - f(\mathcal{A}(x))^2$, where $\mathcal{A}$ is the data augmentation function. Moreover, given value $\xi$, we define $\mathcal{C}_{\mathcal{S}, \chi}(\xi) = \{f \in \mathcal{C}: \widehat{err}_{\rm unl}(f) \leq \xi\}$.
\end{definition}

Then, the following theorems show our main theoretical result, based on the compatibility.

\begin{theorem} \cite{balcan_discriminative_2010} \label{ssl-theorem} Let \mbox{$f^* = \arg\min_{f \in \mathcal{C}_{\phi}} \left[\epsilon(\mathbb{P}, \mathbb{Q}; f) | err_{\rm unl}(f) \leq \xi\right]$}. Then, the following holds, with probability at least $1-\delta$, 
    and for any arbitrarily small $\Delta_{m_{\rm u}, m_{\rm l}}>0$,
    \begin{equation}
    \hat{\epsilon}(S_P, S_Q; f) \leq  \epsilon(\mathbb{P}, \mathbb{Q}; f^*) + \frac{\Delta_{m_{\rm u}, m_{\rm l}}}{2} + \sqrt{\frac{\ln\left(\frac{4}{\delta}\right)}{8m_{\rm u}}}, 
    \end{equation}
    with the unlabelled sample size
    \[
        m_{\rm u} = \mathcal{O}\left(\Delta^{-2} \log \Delta^{-1} \mathcal{V}(\mathcal{C})  + \Delta^{-2}\log(2/\delta)\right),
    \]
    where $\mathcal{V}(\mathcal{C}) = \max\left[VCdim\left(\mathcal{C}\right), VCdim\left(\chi(\mathcal{C})\right)\right]$, and the labelled sample size
    \[
    m_{\rm l} = \frac{8}{\Delta^2}\left[\log\Big(2 \mathcal{C}_{ \mathcal{S}, \chi} (\xi + 2\Delta)\left[2m_{\rm l}, \mathcal{S}\right]\Big) + \log(4/\delta)\right].
    \]
    
    Here, $\chi(\mathcal{C}) = \{\chi_{f} : f \in \mathcal{C}\}$ is assumed to have a finite VC dimension, $\chi_{f}(\cdot) = \chi(f,\cdot)$, and $\mathcal{C}_{\mathcal{S}, \chi}(\xi + 2\Delta)\left[2m_{\rm l}, \mathcal{S}\right]$ is the expected split number for $2m_{\rm l}$ points drawn from $\mathcal{S}$ using functions in $\mathcal{C}_{\mathcal{S}, \chi}(\xi+2\Delta)$. 
\end{theorem}
Theorem \ref{ssl-theorem} indicates that when the best model $f^*$ has an unlabelled error rate of at most $\xi$, the empirical inability of $f$ will be at most $\Delta$ larger than that of $f^*$, with given labelled sample size $m_l$ and unlabeled sample size $m_u$.

\begin{theorem}
    \label{verification_theorem} Let $\mathcal{C} = \{g \circ \phi | \phi \in \mathcal{F}, g \in \mathcal{G} \}$, and suppose $\phi'\in \mathcal{F}$ is fixed (e.g., via pretraining). Then, the following restricted subclass
    \[
       \mathcal{C}_{\phi'}
       \;=\;
       \bigl\{
          g \circ \phi'
          \;\mid\;
          g \in \mathcal{G}
       \bigr\},
       \quad
       \chi(\mathcal{C}_{\phi'})
       \;=\;
       \bigl\{
         \chi_{g\circ \phi'}
         \;\mid\;
         g \in \mathcal{G}
       \bigr\}.
    \]
    satisfy
    \begin{itemize}[leftmargin=1.25em,itemsep=3pt,topsep=4pt]
    \item $\mathcal{C}_{\phi'} \subseteq \mathcal{C}$ 
          and 
          $\chi(\mathcal{C}_{\phi'}) \subseteq \chi(\mathcal{C})$;
    \item $\mathrm{VCdim}\bigl(\mathcal{C}_{\phi'}\bigr) 
           \;\le\; 
           \mathrm{VCdim}\bigl(\mathcal{C}\bigr)$ 
           and 
          $\mathrm{VCdim}\!\bigl(\chi(\mathcal{C}_{\phi'})\bigr) 
           \;\le\; 
           \mathrm{VCdim}\!\bigl(\chi(\mathcal{C})\bigr)$;
    \item $\mathcal{V}(\mathcal{C}_{\phi'}) \le \mathcal{V}(\mathcal{C})$.
    \end{itemize}  
\end{theorem}

\textbf{Interpretations.} Combined with Theorem \ref{ssl-theorem} and Theorem \ref{verification_theorem}, we can find that compared to letting $\phi$ ba learned from scratch, if we learn a fixed $\phi'$ in the representation learning step, we now need fewer unlabeled samples to achieve the same error $\Delta$; or equivalently, given the same unlabeled sample size, we can push $\Delta$ smaller.


\subsection{Proof of Theorem \ref{ssl-theorem}} \label{A2}

\begin{definition}
    Let $\epsilon(\mathbb{P}, \mathbb{Q}; f) \in \left(0, \frac{1}{2}\right)$ be the inability of $f$ to distinguish between distribution $\mathbb{P}$ and $\mathbb{Q}$. Then we define the $err_{\rm te}(f) = 2\epsilon(\mathbb{P}, \mathbb{Q}; f) \in (0, 1)$ to be the error rate of $f$ on distribution $\mathbb{P}$ and $\mathbb{Q}$.
\end{definition}

\begin{theorem} \citep{boucheron_inequality_2000}
    \label{inequality_lemma}
    Suppose function space $\mathcal{C} : \{f | f : \mathcal{X} \rightarrow \{0, 1\}\}$ has finite VC-dimension for $V \geq 1$. For any sample $\mathcal{S}$, any function $f$, we have
    \[
    {\rm Pr}\left[\sup_{f \in \mathcal{C}}|err_{\rm te}(f) - \widehat{err}_{\rm te}(f)| \geq \Delta\right] \leq 8\mathcal{C}[2m_{\rm l}, \mathcal{S}]e^{-m\Delta^2/8}.
    \]
    So for any $\Delta, \delta > 0$, if we draw from $\mathcal{S}$ a sample satisfying 
    \[
    m_{\rm l} \geq \frac{8}{\Delta}\left(\ln(\mathcal{C}[m_{\rm l}, \mathcal{S}]) + \ln\left(\frac{8}{\delta}\right)\right),
    \]
    then, with probability at least $1 - \delta$, all functions $f$ satify $|err_{\rm te}(f) - \widehat{err}_{\rm te}(f)| \leq \Delta$. 
\end{theorem}

\begin{proof}
    The given unlabelled sample size implies that with probability $1 - \delta / 2$, all $f \in \mathcal{C}$ have 
    \[|\widehat{err}_{\rm unl}(f) - err_{\rm unl}(f)| \leq \sqrt{\frac{\ln\left(\frac{4s}{\delta}\right)}{2m_{\rm u}}} \leq \Delta,\]
    which also implies that 
    \[
    \widehat{err}_{\rm unl}(f) \leq err_{\rm unl}(f) + \sqrt{\frac{\ln\left(\frac{4s}{\delta}\right)}{2m_{\rm u}}} \leq \xi + \sqrt{\frac{\ln\left(\frac{4s}{\delta}\right)}{2m_{\rm u}}} \leq \xi + \Delta.
    \]
    Using the standard VC bounds (e.g., Theorem \ref{inequality_lemma}), the labelled sample size $m_{\rm l}$ implies that with probability at least $1 - \delta / 4$, all $f \in \mathcal{C}_{\mathcal{S}, \chi}(\xi + 2\Delta)$ have $|err_{\rm te}(f) - \widehat{err}_{\rm te}(f)| \leq \Delta$.
    Then, by Hoeffding bounds, with probability at least $1 - \delta / 4$ we have 
    \[
    \widehat{err}_{\rm te}(f^*) \leq err_{\rm te}(f^*) + \sqrt{\log(4/\delta)/2m_{\rm l}} \leq err_{\rm te}(f^*) + \Delta.
    \]
    Therefore, with probability at least $1 - \delta$, the $f \in \mathcal{C}$ that optimizes $\widehat{err}_{\rm te}(f)$ subject to $\widehat{err}_{\rm unl}(f) \leq \xi + \Delta$ has 
    \[
    \widehat{err}_{\rm te}(f) \leq err_{\rm te}(f^*) + \sqrt{\frac{\ln\left(\frac{4s}{\delta}\right)}{2m_{\rm u}}} + \sqrt{\log(4/\delta)/2m_{\rm l}} \leq err_{\rm te}(f^*) + \sqrt{\frac{\ln\left(\frac{4s}{\delta}\right)}{2m_{\rm u}}} + \Delta.
    \]
    Moreover, since we have $\widehat{err}_{\rm te}(f) = {\rm Pr}_{(z_i, l_i)\sim\mathcal{S}}\left[f(z_i) \neq l_i\right] \in (0, 1)$ which is proportional to the empirical inability $\hat{\epsilon}(S_P, S_Q; f) \in \left(0, \frac{1}{2}\right)$. Thus, we can conclude the following inequality
    \[
        2\hat{\epsilon}(S_P, S_Q; f) \leq err_{\rm te}(f^*) + \Delta + \sqrt{\frac{\ln\left(\frac{4s}{\delta}\right)}{2m_{\rm u}}},
    \]
    since $err_{\rm te}(f^*) = 2\epsilon(\mathbb{P}, \mathbb{Q}; f^*)$, 
    \[
        \hat{\epsilon}(S_P, S_Q; f) \leq \epsilon(\mathbb{P}, \mathbb{Q}; f^*) + \frac{\Delta}{2} + \sqrt{\frac{\ln\left(\frac{4s}{\delta}\right)}{8m_{\rm u}}},
    \]
    which concludes the proof. 
\end{proof}

\end{document}